# Detecting multiword phrases in mathematical text corpora

*Winfried Gödert*
Cologne University of Applied Sciences
Institute of Information Science

March, 26th 2012

*Abstract*

We present an approach for detecting multiword phrases in mathematical text corpora. The method used is based on characteristic features of mathematical terminology. It makes use of a software tool named *Lingo* which allows to identify words by means of previously defined dictionaries for specific word classes as adjectives, personal names or nouns. The detection of multiword groups is done algorithmically. Possible advantages of the method for indexing and information retrieval and conclusions for applying dictionary-based methods of automatic indexing instead of stemming procedures are discussed.

*Problems and goals*

We start by discussing an example. Given is the text of an abstract for a paper with mathematical content[1]:

> "We study some rigidity properties for locally symmetrical Finsler manifolds. We obtain the local equivalent characterization for a Finsler manifold to be locally symmetric and prove that any locally symmetrical Finsler manifold with nonzero flag curvature must be Riemannian. We also generalize a rigidity result due to Akbar Zadeh."

Looking for methods that will generate index terms automatically and that will have good representation and equally discrimination properties for retrieval purposes, the following question may be of interest:

Which of the words are part of a multiword phrase representing a mathematical concept or a proper entity of mathematical terminology? Intellectual analysis can identify the following phrases:

- rigidity properties
- locally symmetrical Finsler manifold(s)
- local equivalent characterization
- nonzero flag curvature
- rigidity result

We have cited the respective longest sequences with a proper meaning. These sequences can contain shorter ones with normally a generic superordinated meaning.

Next, we ask the following questions. Is it possible to identify sequences by applying automatic techniques? Is it possible to identify as much as possible sequences of words that can be seen as representations of mathematical concepts? Is it possible to avoid identification of almost all sequences that must be seen as senseless or do not have a special mathematical meaning? For example, is it possible to differentiate between the word groups *local equivalent characterization* and *locally symmetrical Finsler manifold* by avoiding the first one for its more general character and to generate the second one? Is it further on possible to identify differently written words like *characterization* or *characterisation* as conceptually equivalent?

---

[1] Abstract taken from the database *Zentralblatt MATH* (http://www.zentralblatt-math.org/zmath/) with permission of the editorial staff.

Purely algorithmic methods of automatic indexing are normally not suited to build clusters of meaningful multiword sequences. They can identify words as character strings with an additional treatment of suffix variations applying stemming methods. Using for example the well-known Porter stemmer for treating the string *locally symmetrical Finsler manifold* leads to the following result[2]:

```
locally -> local
symmetrical -> symmetr
finsler -> finsler
manifolds -> manifold
```

This result does not contain the information that the four words are parts of a conceptual unit. With the aid of a positional index it is possible to identify adjacent words but only if they are formulated in a search string by the searcher. Commonly used procedures cannot distinguish between different word classes like nouns, verbs, adjectives and others. Such a differentiation requires dictionaries with an encoding of word classes. For mathematical text additionally important is a sound identification of proper names used, e.g. special mathematical terms and personal names.

We use a method of automatic indexing that identifies words and word classes on the basis of previously built dictionaries. These dictionaries additionally contain encodings of the suffix behavior of the word classes. Applying the method with these preconditions will show that the answer to our questions mostly will be positive.

*Characteristics of mathematical terminology*

Mathematical text and terminology is characterized by features that are different from other disciplines. We will give a short overview, more detailed discussions can be found in literature (Gödert, 1980; Tietze, 1991; Schwartzman, 1994).

A striking fact for any non-mathematician is the extensive use of words which are quite familiar in everyday language, but with a quite different meaning, for example:

- field, group, ring, tree, wood, sheaf, chain, root, convolution, family, hull, order, trace
- independent, free, normal, entire, ordinary, compact, open, regular, weak, strong, flabby, etc.

This leads to derived terms like *normal family* or *flabby sheaf*, combining an adjective with a noun in order to represent a proper concept. Beside this, there are also words which are used exclusively by mathematicians in a mathematical context, like *diffeomorphism*, *homeomorphism*, *eigenvector*, etc.

For our identification purposes two observations are important. First, it can be observed that a substantial part of mathematical concepts can be described as compounds of adjectives and nouns, e.g.

- distributive algebraic lattice

---

[2] Performed using the Website http://snowball.tartarus.org/demo.php.

- divergent quasilinear parabolic equation
- fourth order nonlinear differential equation
- double-extended quasi-likelihood estimator
- dynamic fourth order partial differential equations
- doubly periodic three-dimensional travelling water waves
- nonlinear parabolic-hyperbolic partial differential equation

Almost any technical terminology of a scientific discipline uses compounds of adjectives and nouns for a generic specification of the nouns' meanings. Comparatively rare, this is done by two or more adjectives as is often the case in the mathematical terminology. The total amount of concepts or subjects of investigation that are represented by multiword sequences is not known. It seems to be far greater than in most other scientific disciplines.

Secondly, it can be observed that many mathematical concepts are named by their inventors in form of so-called *eponyms*:

- Fourier transform
- Poisson distribution
- Lie group

Combining eponyms with adjective-noun-sequences can lead to expressions such as:

- dixmier approximation theorem
- einstein-yang-mills-higgs equations
- einstein-maxwell-gauss-bonnet black hole
- ergodic hamilton-jacobi-bellman equation
- kottler schwarzschild-anti de sitter space-time
- generalized mizoguchi-takahashi's fixed point theorem

Very often such eponyms are formulated in form of adjectives that are derived from the corresponding names, e.g.

- abelian
- smarandachely
- galoisian
- hermitian
- cohopfian
- metabelian
- submetalindelöf

Sometimes we can also find forms of substantiation:

- gaussianity
- gaussians
- gaussoid
- fredholmness
- soberification

It seems desirable to use a technique that can identify the eponym variant *galoisian* by some form of derivation from a dictionary entry *galois* instead of lexicalizing any of these variants. We will see later how this can be done technically.

We can obviously conclude that it might be very helpful for identification purposes to distinguish between words as personal names and words as labels for mathematical concepts. Our example *Lie group* supports this conclusion from an additional point-of-view. The word *lie* is a well-known noun, but in its common meaning not part of mathematical terminology. In a mathematical test it should be identified only as the name of a famous Norwegian mathematician. Realizing this idea requires one or more dictionaries with different encodings for the words, one code for the word *lie* in its common sense meaning and another code for its meaning as personal name[3].

Two cases of homonymy can be observed. Some words have a different meaning in a mathematical context than in everyday situations. Once more the aforementioned practice of different encodings in dictionaries can be used to distinguish the different meanings. Within the context of mathematical texts, this case does not challenge. As we will see later in more detail, one can specify a priority for using different dictionaries. As a result, a word can be identified at first as part of a specialized terminology and only in the case of non-identification as part of everyday language. The second one is an inner-mathematical ambiguity, formed by concepts which occur in different mathematical disciplines, e.g.

| | |
|---|---|
| k-theory | (general topology) |
| | (algebraic topology) |
| | (algebraic geometry) |
| | (commutative rings and algebras) |

It is not possible to offer a trivial solution for disambiguating the different meanings which is based only on the words given.

Mathematical results and text are produced by an international community. The names of their authors are partly written in character sets other than latin. Even if the text source to be analyzed is written in one language - English in the case of our abstracts from the *Zentralblatt MATH* - there may be different spellings of names as a result of different methods for transcribing them from their original language. It should therefore be desirable to identify the variants as synonyms. Similarly, spelling variants of each word can be part of a synonym dictionary (we have indicated this already by our aforementioned example *characterization* or *characterisation*). It should be kept in mind that is not always trivial to decide whether a spelling variant of a name stands for the same or for another person. To achieve homogeneous results, standardizing of personal names should therefore primarily be seen as a problem of the data integrity of the text source and not as much as part of an a posteriori analyzing and indexing procedure.

Further attention should be given to the fact that mathematical text usually contains formulas. Formulas are coded nowadays within a LaTeX-environment, as can be seen by the following example[4]:

> AB: The author finds functions $f: X \to \Bbb R$, $\phi: X \times X \to \Bbb R$ fulfilling the conditions $$\alignat 2 \frac{f(2x + 2y)}{4} - f(x) - f(y) &\geq \phi(x, y),&\quad &x,y \in X,\\ \phi(x, -y) &\geq -\phi(x,y),&\quad &x,y \in X,\\ \limsup_{n \to \infty} \frac{1}{4^{n}} \phi(2^{n}x, 2^{n}x) &< + \infty,&\quad &x \in X,\\ \liminf_{n \to \infty} \frac{1}{4^{n}} \phi(2^{n}x, 2^{n}y) &\geq \phi(x,y),&\quad &x,y \in X, \endalignat $$ where $(X,+)$ is an abelian group. The same is done for a real inner product space $X$ with $ \dim X \geq 3 $ and the conditions $$ \alignat 2 \frac{f(2x + 2y)}{4} - f(x) - f(y) &\geq

---

[3] As part of our investigation, such a dictionary was built up. It contains the names of about 11.500 mathematicians.
[4] Abstract taken from *Zentralblatt MATH*; Zbl 1118.39011.

```
\phi(x, y),&\quad &x,y \in X,\\ \phi(x, -x) &\geq -\phi(x,x),&\quad &x \in X,\\ \phi(2x,2x) &\leq 4\phi(x,x),&\quad &x \in X,\\ \forall_{x \in X} \exists_{n_{0} \in \Bbb N} \forall_{n \geq n_{0}} f\biggl(\frac{x}{2^{n}}\biggr) &\geq 0, &&{}\\ \phi(x,y) &\geq 0, &\quad &x \perp y, x, y \in X .
\endalignat$$.
```

The example shows that there is not much information within the mere text part of this abstract that can be used for our identification method. Thereby, some limits of our method have to be accepted. From this point of view it becomes desirable to use the LaTeX-codes for deriving conceptual content, but this is by no means trivial[5].

Summarizing our analysis of mathematical text so far, we can conclude that it would be very helpful to have a better knowledge about multiword sequences in the mathematical terminology. We will show that the features discussed are the building blocks for an idea of automatically identifying mathematical multiword sequences that can be seen as valuable for describing the content of mathematical text.

*Features of the Lingo indexing system*

The practical part of our investigation was done by using a software tool named *Lingo*. We will give a short overview of its features and show how the tool works in general and how it can be used within our context.

*Lingo* is a linguistic approach to automatic text indexing[6]. It is based on electronic dictionaries and produces index terms. It is an open source system and was designed mainly for research and teaching purposes. The main functions are:

− identification of (i.e. reduction to) basic word form by means of dictionaries and suffix lists,
− algorithmic decomposition,
− dictionary-based synonymization and identification of phrases,
− generic identification of phrases/word sequences based on patterns of word classes.

*Lingo* supports linguistic analysis on text by assembling a network of practically unlimited functionality from a set of modules, called attendees. This network is built by configuration files organizing number and behavior of the attendees. A more complete description of all functions can be found in literature (Lepsky & Vorhauer, 2006; Gödert, Lepsky & Nagelschmidt, 2011, especially part 5). Generally two different modes for processing text can be chosen. One mode can be used to generate index entries for retrieval purposes if the text file is logically organized in form of data records. The other mode is suitable for analyzing a text file as a whole to produce collections of words which can be used for building up dictionaries or for deriving distribution properties of index terms.

*Lingo* starts indexing of a data source by reading the text sequentially. The file is read and its content is put into the channels line by line. On the basis of a rule set, an attendee called *tokenizer* dissects lines into defined character strings, called *tokens*. By hypothesis these tokens are proposed as candidates for words. In a next step these candidates are checked against dictionaries by an attendee called *wordsearcher*. If there are two or more dictionaries to be checked a priority order has to be defined. If there is a match the token tested is qualified as a word, otherwise it is written into a protocol file of tokens not recognized. The protocol files can be analyzed subsequently for enhancing the dictionaries by new words.

---

[5] Some research on this problem is in progress; first reports have been given (Sperber & Wegner, 2011; Sperber & Ion, 2011).
[6] Cf. http://www.lex-lingo.de. A simple Web interface for a first impression can be found under: http://linux2.fbi.fh-koeln.de/lingoweb/.

The use of more than one dictionary is the standard case for applying a system dictionary and one or more user dictionaries. System dictionaries are updated whenever a new version of *Lingo* is installed.

It can be seen as an important feature of *Lingo* that the identification is not only based on the word pattern as specified in the dictionary. The identification is based also on additional suffixes from a parameter file. For example, a noun is identified if it appears in its basic form or with additional suffixes like

   es s ves/f ves/fe ies/y

For example, the plural word *houses* is identified as a valid word on basis of a dictionary entry *house* and adding the suffix *s*. The specifications with a slash are used for substitutions in cases like *families - family*. The resulting index term is always given in its basic form as specified in the dictionary.

Suffixes can be specified for each word class considered as necessary or useful. Typical examples for word classes are nouns, verbs, adjectives, proper names for persons and institutions or high-frequency words.

The results of the attendee *wordsearcher* form the basis for the work of all subsequent attendees. An attendee named *synonymer* extends words with synonyms. Synonyms can be true synonyms in the sense of spelling variants or conceptually equivalent words but it is also permissible to include near-synonyms. Indexing synonyms or near-synonyms goes beyond the inclusion of suffix variants for there is no need that the word patterns must have something in common. The decision about synonymy of two words is specified by a separate dictionary.

An attendee named *decomposer* has been developed for the treatment of compound terms (*Komposita*) occurring very often in German texts[7]. The *decomposer* tests any token not identified by the *wordsearcher* for being a compound. The decision whether or not a token is a compound is based on occurrence of the respective parts as entries in dictionaries. The compound and its derived parts can be used as index terms. In this way the retrievable vocabulary is substantially enhanced in order to improve recall.

Two attendees can be used for treating multiword phrases, the *multiworder* and the *sequencer*. The *multiworder* identifies phrases (word sequences) based on a multiword dictionary. It has high potential for precise identification of sequences with a conceptual quality. To meet the specific terminological requirements of the text to be indexed, the dictionary has previously to be defined. The second attendee for identifying multiword sequences, the so-called *sequencer* works purely algorithmic. It is based on the identification results of the *wordsearcher* and uses additionally predefined patterns of word classes to identify sequences. For example, the pattern *AAS* represents a sequence of two subsequent adjectives and one noun. Any respective sequence including suffix and synonym variants will be identified. Since this attendee plays an important part in our investigation it will be explained in more detail in the next section.

---

[7] We do not make use of this feature. Decomposition is a central part of the software for treating text in German, a language in which composites form substantial part of the lexical units. For a closer description of this feature we refer to literature (Lepsky & Vorhauer, 2006; Gödert, Lepsky & Nagelschmidt, 2011, especially part 5).

There are some additional attendees like the *abbreviator*, *variator* or *dehyphenizer*. They were not used within the context of our investigation. For a detailed description we refer to the *Lingo* documentation[8].

*Outline of the method for identifying multiword sequences*

Our approach of identifying multiword sequences by the *Lingo* attendee *sequencer* requires proceeding by several steps. First of all, the data source has to be reduced to mere text strings by eliminating all LaTeX-encodings. This step is not necessary for obtaining results but it is beneficial for the performance and for reducing the size of the file containing the unidentified tokens. The resulting "reduced" file has to be analyzed to select words that can be seen as parts of mathematical terminology. For our investigations these words were arranged in three different dictionaries defining respective word classes:

− one dictionary of adjectives in the scope of approximately 10.000 entries,
− one dictionary of proper names (nouns) of mathematical terminology in the scope of approximately 13.000 entries,
− one dictionary of personal names of mathematicians in the scope of approximately 11.500 entries.

Comparing these figures with the number of entries in the *Encyclopedia of Mathematics*[9] (EoM) or the *Mathematics Subject Classification* (MSC2010) they may not seem very impressive. But as we have already noticed, many mathematical concepts are formulated as multiword phrases for which our dictionaries contain the combinatorial components. Later on we will get an impression of how many subjects can be generated by the words of our dictionaries. Not all entries can be seen as elements of mathematical terminology in a strong sense but they all occurred in a database of mathematical literature, so they were considered relevant for our dictionaries.

For identifying a token in the dictionaries by the attendee *wordsearcher* it is necessary to specify a procedural order. We have chosen the following one:

1. adjectives
2. proper names
3. personal names
4. system dictionary

The main idea of this order is to prefer words which are closer to mathematical terminology. Accordingly, we prefer the words contained in our dictionaries of mathematical adjectives and proper names instead to favor the more general words that are contained in the system dictionary. If a token is identified as a word in one dictionary it is not checked against a subsequently ordered dictionary. It is left to experience whether the order used for our investigations is the best one.

For each of the word classes a set of suffixes was defined which allows identification of respective word variants. For example, character strings such as *abelian* or *galoisian* are typical adjective similar variations of personal names frequently used in mathematical texts. They can now be identified by the dictionary entries *abel* or *galois* and the additional suffix *-ian* defined for the word class of personal names. Similarly, character strings like *gaussianity*, *gaussians* or *gaussoid* are identifiable by the dictionary entry *gauss*.

---

[8] Cf. http://www.lex-lingo.de.
[9] The 2002 version of the *Encyclopedia of Mathematics* contains more than 8,000 entries and 50.000 notions.

By the dictionaries and suffix definitions the essential conditions are given for identifying words and their variants that can be seen as parts of mathematical terminology and equally as parts of word sequences. In a next step patterns of word classes had to be defined that seemed to be suitable for identifying good word sequences and to avoid uninteresting ones. To construct a pattern list, we allocated for a test set the respective word class indicators. We show some examples in the following list:

- distributive algebraic lattice - AAS
- nonlinear schrödinger equation - ANE
- algebraic riccati equation - ANE
- divergent quasilinear parabolic equation - AAAE
- stochastic partial differential equation - AAAE
- dynamic fourth order partial differential equations - AAEAAE
- incompressible navier stokes equation - ANNE
- ergodic hamilton-jacobi-bellman equation - ANNNE
- krasnoselskii fixed point theorem - NAEE
- doubly quasi-periodic riemann boundary value problem - AAANEEE
- einstein-yang-mills-higgs equations - NNNNE
- einstein-maxwell-gauss-bonnet black hole - NNNNAS
- generalized mizoguchi-takahashi's fixed point theorem - ANNAEE

The resulting first set of sequences was enhanced by combinatorial considerations to derive a set for the final investigation. The longest pattern used consisted of 8 words. The complete list is shown below:

AAAAAE, AAAAAEE, AAAAAES, AAAAAS, AAAAE, AAAAEAE, AAAAEAS, AAAAEE, AAAAEEE, AAAAEES, AAAAES, AAAAS, AAAASE, AAAE, AAAEE, AAAEEE, AAAEEEE, AAAEEES, AAAEES, AAAES, AAANEEE, AAANEES, AAANESS, AAANNNE, AAANNNS, AAANSEE, AAANSES, AAANSSE, AAANSSS, AAAS, AAEAAE, AAEAAES, AAEAAS, AAEAE, AAEAES, AAEANE, AAEANS, AAEAS, AAEE, AAEEAES, AAEEAS, AAEEE, AAEES, AAES, AAESS, AANAE, AANAS, AANE, AANEE, AANEEE, AANEES, AANES, AANESS, AANNAE, AANNAES, AANNAS, AANNE, AANNES, AANNNAE, AANNNAS, AANNNE, AANNNNAE, AANNNNAS, AANNNNE, AANNNNS, AANNNS, AANNS, AANS, AANSEE, AANSES, AANSSE, AANSSS, AAS, AASAAS, AASAES, AASANE, AASANS, AASAS, AEAAAE, AEAAAS, AEAAE, AEAAES, AEAAS, AEAS, AEEAE, AEEAES, AEEEAS, AEEEES, AEENEEE, AEES, AEESS, AENEE, AENES, AES, AESNEEE, AESNEES, AESS, ANE, ANEEE, ANEES, ANNAE, ANNAEE, ANNAES, ANNAS, ANNE, ANNES, ANNNAE, ANNNAS, ANNNE, ANNNNAE, ANNNNAS, ANNNNE, ANNNNS, ANNNS, ANNS, ANS, AS, ASAAE, ASAAES, ASAAS, ASAE, ASAES, ASAS, ASS, EAAAE, EAAAES, EAAAS, EAAS, EAS, EE, EEAAS, EEAE, EEAES, EEAS, EEE, EEEE, EEES, EES, EESS, ES, ESS, NAAAE, NAAAEE, NAAAES, NAAAS, NAAE, NAAS, NAE, NAEE, NAES, NAS, NE, NNAE, NNAEE, NNAES, NNANNEE, NNANNES, NNAS, NNE, NNEEAE, NNEEAS, NNES, NNESAE, NNESAS, NNNAE, NNNAS, NNNE, NNNNAE, NNNNAS, NNNNE, NNNNEE, NNNNES, NNNNNAE, NNNNS, NNNS, NNS, NS

Although lengthy, this list does not contain all possible combinations of 2 to 8 words including 4 word classes. For performance reasons, we have chosen only those that seem to be most plausible for formulating mathematical concepts by multiword sequences. It must be left to experience whether this list is almost complete or deficient.

For a more detailed impression how texts are being treated by *Lingo* we give an extract of the protocol file for the example *locally symmetrical Finsler manifold* already mentioned at the beginning of our discussion. The treatment of this string reads as follows:

    lex:)  <locally = [(local/a)]>
    lex:)  <symmetrical = [(symmetric/a)]>
    lex:)  <Finsler = [(finsler/n)]>
    lex:)  <manifolds = [(manifold/e)]>

All words could be identified by a dictionary entry, otherwise they would have been marked by a *?* instead of assigning an indicator for a word class. The brackets contain the respective results of the identification, the word checked is written as its dictionary entry supplemented by a code for the word class. The next three lines show the results of the pattern identification by the attendee *sequencer*.

    lex:)  <finsler manifold|SEQ = [(finsler manifold/q)]>
    lex:)  <symmetric finsler manifold|SEQ = [(symmetric finsler manifold/q)]>
    lex:)  <local symmetric finsler manifold|SEQ = [(local symmetric finsler manifold/q)]>

The results of identifying tokens as the single words are similar to the results a stemming procedure such as the Porter stemmer would generate. In fact, there is a difference. *Lingo* does not produce word stems but root words. The basic improvement is shown in the three lines containing sequences. These lines show the results of the word class based identification process of the *sequencer* for the patterns *NE*, *ANE* and *AANE*. For generating this result the dictionary to be consulted must contain a form of knowledge about the corresponding word classes by encodings.

### *Results of identifying multiword sequences*

We now present some results of our sequence identification for the more than 700.000 documents of our data source originating from the database *Zentralblatt MATH*. For our investigation we only used the words of the abstracts. We excluded additionally existent keywords - mostly but not always multiword phrases - that are indexed by the editorial staff. The main reason for this decision was the idea that the identification procedure should work as close as possible only on a given text and not on word material intellectually produced or refined.

In total, 1.722.711 different sequences were identified. We will present some results from a statistical point of view.

First, we take a look at the top 50 sequences with highest frequency:

    16457  boundary condition
    14138  second order
    12926  finite element
    12890  boundary value
    12305  banach space
    11004  result show
     9646  lower bound
     9143  first order
     9029  optimization problem
     9001  upper bound
     8535  hilbert space
     7780  paper deal

```
7300  dynamical system
7255  lie algebra
6892  volume will
5916  integral equation
5871  higher order
5369  simulation result
5323  black hole
5278  linear system
5132  point theorem
4967  data set
4952  continuous function
4896  control problem
4762  heat transfer
4691  time dependent
4503  lie group
4498  time series
4385  space time
4369  field theory
4257  vector space
4208  schrödinger equation
4131  markov chain
4114  riemann manifold
4098  stokes equation
3924  review copy
3812  objective function
3736  cauchy problem
3725  navier stokes equation (*)
3714  decision making
3676  control system
3656  difference equation
3632  time delay
3614  wave equation
3600  initial value
3582  boundary layer
3565  paper study
3562  previous work
3539  state space
3535  reynolds number
```

Only one sequence with 3 parts occurs in the top 50 (*navier stokes equation*, marked by (*)), the rest consists of 2 parts.

The distribution of sequences with n parts is as follows:

```
8 parts:          1  time
7 parts:      2.164  times
6 parts:     21.392  times
5 parts:    117.170  times
4 parts:    392.032  times
3 parts:    573.511  times
2 parts:    616.417  times
```

For sequences with a defined number of parts we show the beginning of the resulting lists representing the sequences with highest frequencies. Within the excerpts presented here, we did not make any omissions:

| | | | |
|---|---:|---|---|
| 8 parts: | 1 | time | well known chen harker kanzov smale smooth function |
| 7 parts: | 7 | times | strong nonlinear mixed variational like inequality problem |
| | 6 | times | symmetric informational complete positive operator valued measure |
| | 5 | times | fully three dimensional finite strain damage model |
| 6 parts: | 27 | times | finite generated torsion free nilpotent group |
| | 19 | times | strong nonlinear mixed variational like inequality |
| | 18 | times | necessary as well as sufficient condition |
| | 18 | times | one dimensional backward stochastic differential equation |
| | 14 | times | finite generated fully residual free group |
| 5 parts: | 288 | times | second order ordinary differential equation |
| | 124 | times | second order partial differential equation |
| | 114 | times | fully polynomial time approximation scheme |
| | 108 | times | first order partial differential equation |
| | 105 | times | leggett williams fixed point theorem |
| 4 parts: | 948 | times | published without additional control |
| | 882 | times | nonlinear partial differential equation |
| | 591 | times | second order differential equation |
| | 581 | times | two point boundary value |
| | 499 | times | nonlinear ordinary differential equation |
| | 496 | times | common fixed point theorem |
| | 482 | times | incompressible navier stokes equation |
| | 472 | times | schauder fixed point theorem |
| 3 parts: | 3725 | times | navier stokes equation |
| | 1900 | times | initial boundary value |
| | 1849 | times | experimental result show |
| | 1657 | times | simulation result show |
| | 1443 | times | closed loop system |
| | 1344 | times | quantum field theory |
| | 1273 | times | numerical result show |
| | 1265 | times | support vector machine |
| | 1194 | times | point boundary value |
| | 1059 | times | probability density function |

One will immediately recognize that the excerpts contain sequences not corresponding to mathematical terminology. Regarding the sequence *well known chen harker kanzov smale smooth function* one can argue that the two words *well known* do not matter, the sequence without both words is produced additionally by our identification patterns. In fact, the shorter sequence occurs 1 time, but it could have occurred more frequently. The sequence *published without additional control* might be of interest for studying publishing procedures, obviously it is not of interest for mathematical questions. It depends on the point of view whether its occurrence should be seen as a severe problem. We think that this is not so, such phenomena cannot be avoided by the method presented so far, it can only be influenced by optimizing the dictionaries, the suffixes and the identification patterns for the sequences. To achieve further improvements more experience in applying the method or using another form of dictionary is necessary. We will describe such a procedure at the end of the present section.

The next list shows how many different sequences - independent of their length - could be found occurring n times:

```
n=1:      1.241.515  sequences
n=2:        203.522  sequences
n=3:         80.353  sequences
n=4:         44.057  sequences
n=5:         27.557  sequences
n=10:         6.854  sequences
n=15:         3.175  sequences
n=20:         1.764  sequences
n=50:           278  sequences
n=100:           74  sequences
n=200:           18  sequences
n=300:           10  sequences
n=400:            4  sequences
n=500:            2  sequences
```

In a graphical visualization the total distribution looks as is shown in Fig. 1.

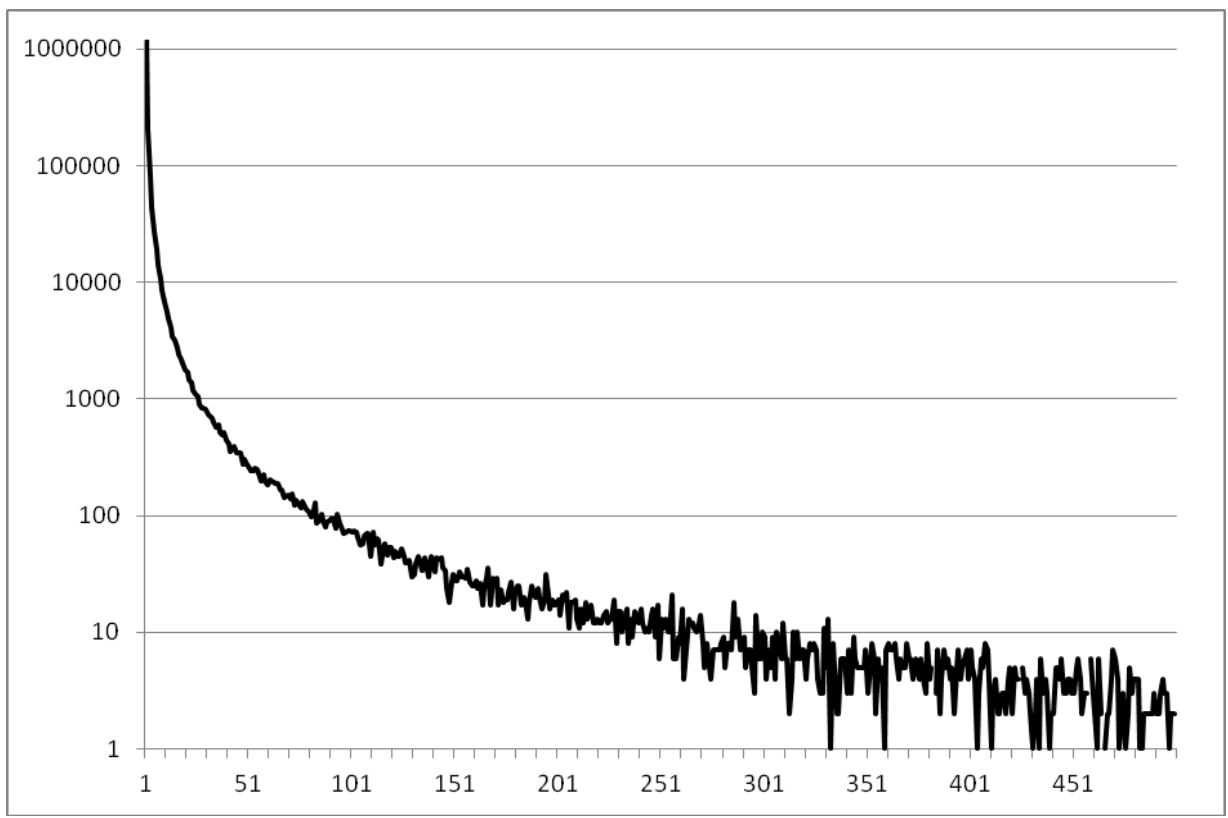

Fig. 1: Frequency of sequences (1 to 500 times of occurrence) in a logarithmic scale

Fig. 2 and Fig. 3 show a closer look at two segments of the total distribution

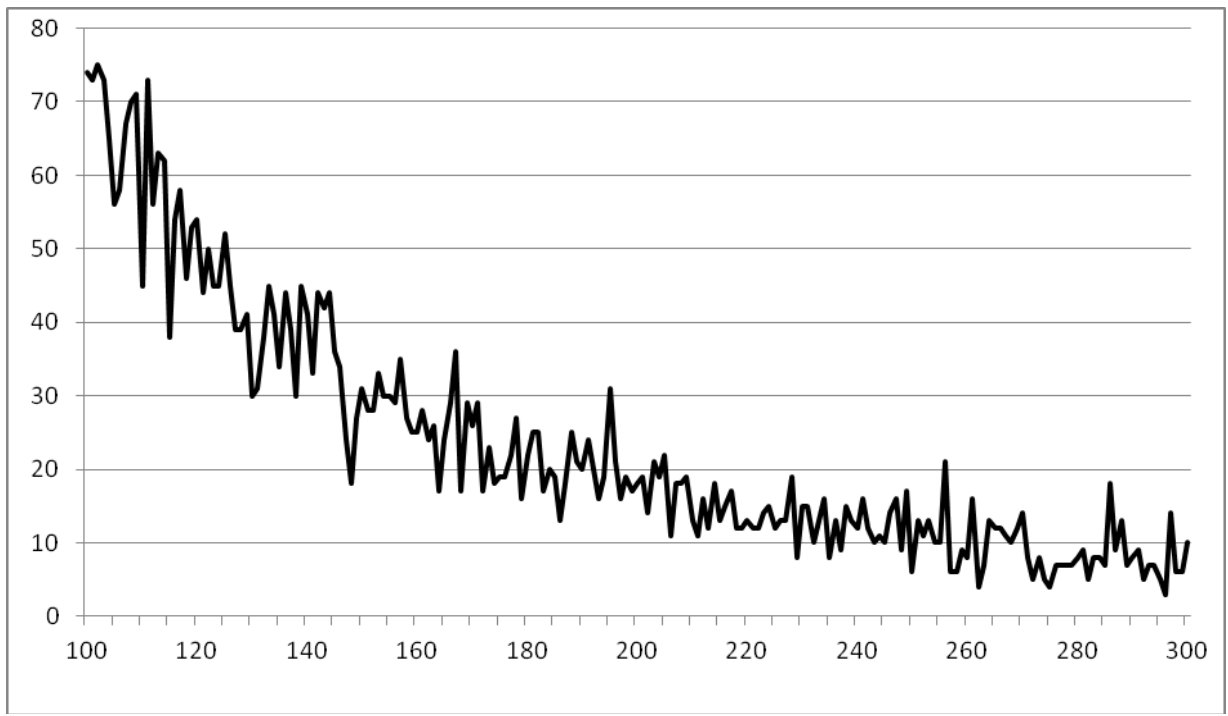

Fig. 2: Frequency of sequences (100 to 300 times of occurrence)

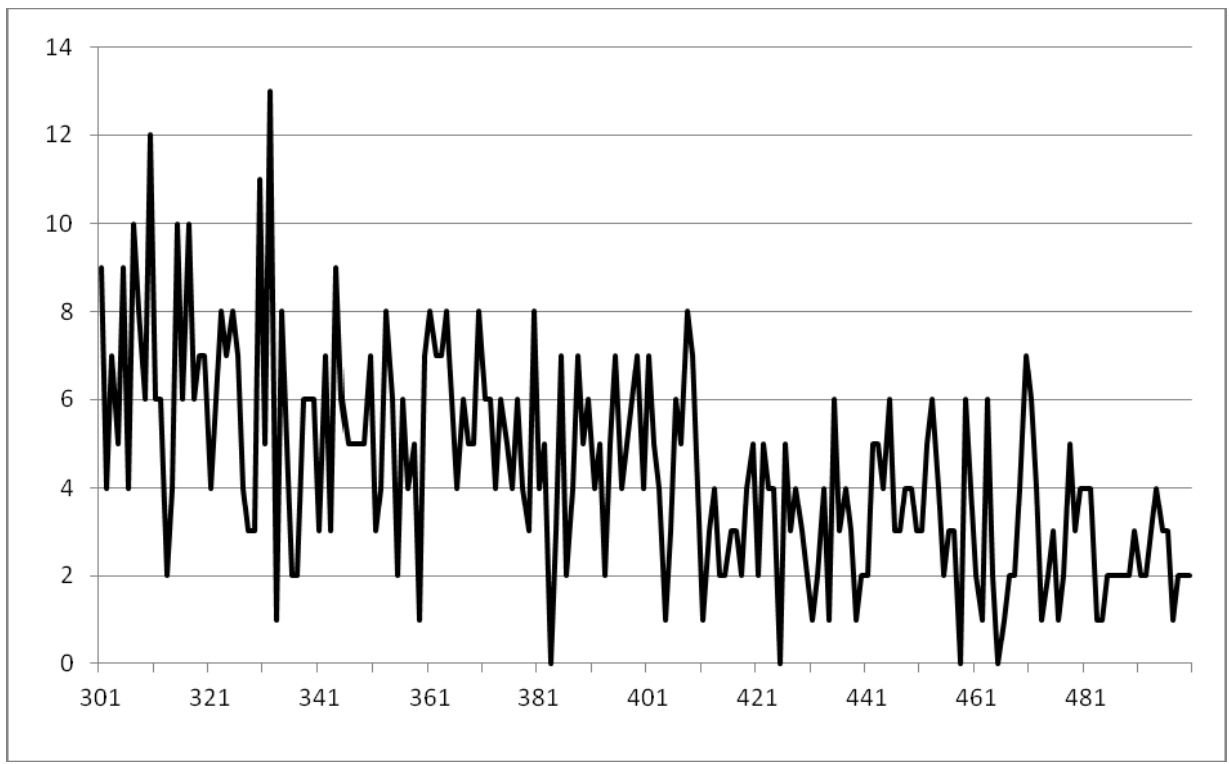

Fig. 3: Frequency of sequences (301 to 500 times of occurrence)

To give a condensed overview of the sequences' frequencies in correlation with the number of their parts we have clustered the data as shown in the following summaries:

*Sequences with frequency > 500 times*

| Total | 5 or more parts | 4 parts | 3 parts | 2 parts |
|---|---|---|---|---|
| 902 | 0 | 4 | 47 | 851 |

*Sequences with frequency > 200 times*

| Total | 6 or more parts | 5 parts | 4 parts | 3 parts | 2 parts |
|---|---|---|---|---|---|
| 2.917 | 0 | 1 | 27 | 209 | 2.680 |

*Sequences with frequency > 100 times*

| Total | 6 or more parts | 5 parts | 4 parts | 3 parts | 2 parts |
|---|---|---|---|---|---|
| 6.428 | 0 | 6 | 72 | 569 | 5.781 |

*Sequences with frequency > 50 times*

| Total | 6 or more parts | 5 parts | 4 parts | 3 parts | 2 parts |
|---|---|---|---|---|---|
| 13.501 | 0 | 19 | 257 | 1.412 | 11.813 |

*Sequences with frequency > 20 times*

| Total | 6 or more parts | 5 parts | 4 parts | 3 parts | 2 parts |
|---|---|---|---|---|---|
| 34.171 | 1 | 98 | 990 | 4.671 | 28.411 |

In the next section we take a closer look on those groups that contain one or more personal names. We have identified 267.280 such sequences. This equals 15,5 percent of our totally detected 1.722.711 sequences. The top 50 of this list reads[10]:

| | |
|---:|---|
| 12305 | banach space |
| 8535 | hilbert space |
| 7255 | lie algebra |
| 4503 | lie group |
| 4208 | schrödinger equation |
| 4131 | markov chain |
| 4114 | riemann manifold |
| 4098 | stokes equation |
| 3736 | cauchy problem |
| 3725 | navier stokes equation |
| 3535 | reynolds number |
| 3524 | fourier transform |
| 2976 | green function |

---

[10] We have deleted 7 sequences from the original list - black hole (5323), cross section (2553), case study (2528), good agreement (2259) white noise (1317), he show (1194), low frequency (930) - that contains words encoded as personal names and as nouns.

| | |
|---:|:---|
| 2466 | sobolev space |
| 2409 | hopf algebra |
| 2257 | lyapunov function |
| 2118 | dirichlet boundary |
| 2079 | riemann surface |
| 1897 | hopf bifurcation |
| 1857 | newton method |
| 1836 | laplace transform |
| 1689 | markov process |
| 1644 | galerkin method |
| 1624 | nonlinear schrödinger equation |
| 1575 | banach algebra |
| 1566 | dirichlet problem |
| 1512 | lyapunov exponent |
| 1445 | euler equation |
| 1423 | petri net |
| 1413 | kähler manifold |
| 1396 | schrödinger operator |
| 1366 | fourier series |
| 1348 | hardy space |
| 1342 | maxwell equation |
| 1293 | nash equilibrium |
| 1273 | lebesgue measure |
| 1247 | levy process |
| 1230 | hausdorff dimension |
| 1186 | neumann boundary |
| 1174 | poisson process |
| 1116 | galois group |
| 1079 | dirac operator |
| 1067 | lagrange multiplier |
| 1057 | poisson equation |
| 1053 | neumann algebra |
| 1052 | boltzmann equation |
| 1023 | burgers equation |
| 969 | gordon equation |
| 947 | riccati equation |
| 913 | nusselt number |

The total number of sequences that contain personal names in correlation to the number of parts without regarding their frequency is given in the following overview:

| Total | 8 parts | 7 parts | 6 parts | 5 parts | 4 parts | 3 parts | 2 parts |
|---:|---:|---:|---:|---:|---:|---:|---:|
| 267.280 | 1 | 150 | 1.869 | 13.690 | 50.606 | 112.280 | 88.684 |

In more detail, we can present the following relative frequency numbers in correlation with number of parts:

*Sequences with frequency > 500 times*

| Total | 5 or more parts | 4 parts | 3 parts | 2 parts |
|-------|-----------------|---------|---------|---------|
| 141   | 0               | 0       | 16      | 125     |

*Sequences with frequency > 200 times*

| Total | 5 or more parts | 4 parts | 3 parts | 2 parts |
|-------|-----------------|---------|---------|---------|
| 478   | 0               | 9       | 78      | 391     |

*Sequences with frequency > 100 times*

| Total | 5 or more parts | 4 parts | 3 parts | 2 parts |
|-------|-----------------|---------|---------|---------|
| 1.082 | 1               | 24      | 200     | 857     |

*Sequences with frequency > 50 times*

| Total | 5 or more parts | 4 parts | 3 parts | 2 parts |
|-------|-----------------|---------|---------|---------|
| 2.345 | 5               | 75      | 478     | 1.787   |

*Sequences with frequency > 20 times*

| Total | 5 or more parts | 4 parts | 3 parts | 2 parts |
|-------|-----------------|---------|---------|---------|
| 6.322 | 17              | 244     | 1.568   | 4.493   |

For a closer impression we present lists of the higher frequent sequences with personal names in correspondence to the number of their parts:

*Sequences with 3 parts occurring more than 500 times:*

  navier stokes equation
  nonlinear schrödinger equation
  runge kutta method
  separable hilbert space
  real banach space
  klein gordon equation
  compact riemann manifold
  riemann zeta function
  hamilton jacobi equation
  large eddy simulation
  fokker planck equation
  reflexive banach space
  yang mills theory
  euler lagrange equation
  schauder fixed point
  complex hilbert space

*Sequences with 4 parts occurring more than 200 times:*

   incompressible navier stokes equation
   schauder fixed point theorem
   uniform convex banach space
   banach fixed point theorem
   uniform smooth banach space
   infinite dimensional hilbert space
   compressible navier stokes equation
   hamilton jacobi bellman equation
   infinite dimensional banach space

*Sequences with 5 parts occurring more than 20 times:*

   leggett williams fixed point theorem
   three dimensional navier stokes equation
   leray schauder fixed point theorem
   two dimensional navier stokes equation
   generalized hyers ulam rassias stability
   dimensional incompressible navier stokes equation
   partial observable markov decision process
   self dual yang mills equation
   continuous time markov decision process
   calderon zygmund singular integral operator
   *ha only finite many solution (\*)*
   local convex hausdorff topological vector
   affine kac moody lie algebra
   dimensional compressible navier stokes equation
   new hilbert type integral inequality
   fan glicksberg fixed point theorem
   unsteady incompressible navier stokes equation

*Sequences with 6 parts occurring more than 5 times:*

   one dimensional time independent schrödinger equation
   known leggett williams fixed point theorem
   well known leggett williams fixed point
   hausdorff local convex topological vector space
   ha infinite many positive integer solution
   simply connected compact simple lie group
   five dimensional myers perry black hole

It should again be noted that the result of our identification method contains sequences that cannot be seen as proper mathematical subjects. We have already mentioned the example *published without additional control*. For sequences containing eponyms we can give the additional example *ha only finite many solution* (in our list of sequences with 5 parts marked by (\*)). This sequence contains the token *ha*, encoded as a personal name. In fact, our result must contain examples like this as long as the identification is based on a simple algorithmic identification of word classes. A substantial improvement is possible using an additional dictionary of multiword groups containing intellectually verified sequences. The attendee *multiworder* will identify only those sequences as word groups that are contained in this dictionary and will add them as index terms to the data records. We have done this for a smaller part of our document base with orientation to the subject fields *graph theory* and *ordinary differential equations*. A first evaluation has shown, that this method yields very good results suitable at least for a recommender system supporting intellectual indexing. But this only is valid under two prerequisites. At first, the proper multiword groups have to be selected from the sequencer's result to build the multiword dictionary, a laborious task. Secondly, there should be a continuous enhancement of the multiword dictionary with new sequences that are results of an identification process by the *sequencer*. It seems as if nowadays there is a much more dynamical process of creating new sequences within the mathematical community than in former days. A respective workflow would have to be defined for a real production system.

*Perspectives*

Concluding the report of our investigations we we want to discuss some benefits of our method, which give rise to further improvements.

A first benefit arises from the fact that the sequences can be used as index terms for data records. This has already been mentioned. If no automatic assignment is preferred, a recommender process can be established that allows the indexer to decide whether to use the results or not.

A second benefit can be seen in the opportunity to get an overview of the multiword phrases used in a specialized subject discipline or context. If furthermore the assumption is valid that a substantial part of the disciplines' concepts are represented by multiword phrases, one gets insight into actual concepts and themes of the discipline. This overview might be more comprehensive and perhaps even better than structured compilations like encyclopedias or indexing languages. For example, we can compare the following figures with the number of our sequences. The *Encyclopedia of Mathematics* (EoM) contains 8.000 entries and 50.000 notions,  the *Mathematics Subject Classification* (MSC2010) contains 6.000 classes. If the assumption is justified that a sequence that is used more than 20 times represents an accepted concept - and not only an idiosyncratic one - our figures show a much finer granularity of conceptual discrimination. There are 6.428 sequences with frequency more than 100 times, 13.501 sequences with frequency more than 50 times or 34.171 sequences with frequency more than 20 times. This insight can at first be used for enhancing and improving mathematical indexing languages like classification systems, thesauri or even more elaborated conceptual structures as semantic networks or ontologies. Tools like that can be further beneficial for designing navigational aids as interfaces for retrieval systems.

A third benefit should be object for further research. The identified words and sequences in their linguistic standardized form can be used for statistical analysis. It can be assumed that the normalization will yield better results for generating conceptual clusters than the use of words or sequences without normalization. Consequently, one would predict better results for all methods of information retrieval based on statistical results of the counting of words or sequences. Examples may include ranking algorithms or clustering techniques such as Latent Semantic Indexing.

The proposed method can potentially be improved through the analysis of sequences containing prepositions. This word class is usually ignored because prepositions are considered only as high frequent words with no independent significance for content representation. If seen in context with multiword phrases this opinion might be worth to correct. We give some examples:

- drinfeld-jimbo quantization of lie bialgebra
- elliptic system of fitzhugh-nagumo type
- nonlinear boundary value problems for ordinary differential equations

- nonlinear deterministic systems with complex dynamical behavior
- nonlinear parabolic boundary value problems with singular coefficients
- duffing equation with an integral term of non-viscous damping
- fuzzy volterra integro-differential equations of the second kind
- first order nonlinear impulsive integro-differential equation of volterra type
- entire functions of completely regular growth in the sense of levin and pfluger
- essential self-adjointness of spatial part of reduced normalized wave operator of kerr metric
- exact solutions of the compound korteweg-de vries-sawada-kotera equation
- fourier transform of a probability measure on a locally compact group at a unitary representation

If prepositions are to be considered for an analysis it is necessary to build up a special dictionary for prepositions and to formulate an extended set of word class patterns. The examples given make it immediately clear that this set of patterns will be more extensive and complex than the one used for our present investigations. It is left to further research to study this idea more closely.

Our investigations and observations were focused exclusively on text corpora with mathematical content. Yet, it might be interesting to apply the method presented here for identifying multiword sequences in texts of other scientific disciplines. Doing so, one can find out whether and in which degree comparable results occur.

Finally, we can conclude that our investigation gives support for applying a dictionary-based method of automatic indexing for subject specific texts in English language, a domain commonly more seen as an application field for stemming procedures.

*Acknowledgements*

Part of this work was done in cooperation with *Zentralblatt MATH*. The work would not have been possible without the data records of more than 700.000 documents that were provided by the editorial staff. Special thanks to *Wolfram Sperber* for valuable comments and hints. Furthermore I have to thank my colleagues *Klaus Lepsky* for tirelessly giving stimulating comments and *Jens Wille* for continuously supporting the procedural part of indexing the text files and developing numerous improvements for the software tool *Lingo*.